\documentclass[10pt,twocolumn,a4paper,fleqn]{article}
\usepackage[a4paper,left=15mm,right=15mm,top=30mm,bottom=25mm,columnsep=10mm]{geometry}
\usepackage{graphicx}
\usepackage{times}
\usepackage{titlesec}
\usepackage{indentfirst}
\usepackage[fleqn]{amsmath}
\usepackage{fancyhdr}
\usepackage{cite}
\usepackage{enumitem}
\usepackage{etoolbox}

\renewcommand{\normalsize}{\fontsize{10pt}{11.9pt}\selectfont} 
\titleformat{\section}
  {\normalfont\large\bfseries\MakeUppercase}{\thesection.}{0.5em}{}
\titlespacing*{\section}{0pt}{2.5ex plus 0.5ex minus .2ex}{1.5ex}

\titleformat{\subsection}
  {\normalfont\normalsize\bfseries}{\thesubsection.}{0.5em}{}
\titlespacing*{\subsection}{0pt}{1.5ex}{0.5ex}

\titleformat{\subsubsection}
  {\normalfont\normalsize}{\thesubsubsection.}{0.5em}{}
\titlespacing*{\subsubsection}{0pt}{2.0ex plus .5ex minus .2ex}{0.1ex}

\setlist[description]{
  font=\itshape,
  nosep,
  leftmargin=0pt,
  labelsep=0.5em
}

\AtBeginEnvironment{table}{
  \scriptsize
}

\usepackage[small,bf]{caption}
\DeclareCaptionLabelSeparator{periodspace}{.~ }
\makeatletter
\newcommand{\affils}[1]{\def\@affils{#1}}
\renewcommand{\abstract}[1]{\def\@abstract{#1}}
\newcommand{\keywords}[1]{\def\@keywords{#1}}

\renewcommand{\@maketitle}{%
  \newpage
  \null
  \begin{center}
    {\fontsize{15pt}{15pt}\selectfont \bfseries \@title \par}
    \vskip 1.0em
    {\large \@author \par}
    \vskip 0.5em
    {\normalsize \@affils \par}
  \end{center}
  {\normalsize \noindent \textbf{Abstract:} \@abstract \par}
  \vskip 1em
  {\normalsize \noindent \textbf{Keywords:} \@keywords \par}
  \vskip 1.4em
}
\makeatother

\makeatletter
\renewenvironment{thebibliography}[1]{
  \section*{\refname}
  \normalsize
  \list{[\arabic{enumi}]}{
    \settowidth\labelwidth{[#1]}
    \leftmargin\labelwidth
    \advance\leftmargin\labelsep
    \setlength{\itemsep}{0pt}
    \setlength{\parsep}{0pt}
    \setlength{\topsep}{0pt}
    \setlength{\partopsep}{0pt}
    \usecounter{enumi}
    
  }
  \def\newblock{\hskip .11em plus .33em minus .07em}
  \sloppy\clubpenalty4000\widowpenalty4000
  \sfcode`\.=1000\relax
}{
  \endlist
}
\makeatother

\usepackage{amssymb}
\usepackage{algorithmicx}
\usepackage{algorithm}
\usepackage{algpseudocodex}
\usepackage{bm}
\renewcommand{\boldsymbol}[1]{\bm{#1}}

\graphicspath{{content/}}

\fancypagestyle{firstpage}{
  \fancyhf{}
  
  \fancyfoot[C]{\footnotesize \sffamily Author's version. In Proc. of the Joint Symposium of AROB 31st and ISBC 11th (AROB-ISBC 2026), pp.~1615--1619, 2026.}
}

\title{Data-Driven Control of a Magnetically Actuated Fish-Like Robot}

\author{Akiyuki Koyama${}^{1}$ and Hiroaki Kawashima${}^{1}$}

\affils{${}^{1}$ Graduate School of Information Science, University of Hyogo, Japan\\
(E-mail: akiyuki.koyama@gmail.com, kawashima@gsis.u-hyogo.ac.jp)\\
}

\abstract{%
Magnetically actuated fish-like robots offer promising solutions for underwater exploration due to their miniaturization and agility; however, precise control remains a significant challenge because of nonlinear fluid dynamics, flexible fin hysteresis, and the variable-duration control steps inherent to the actuation mechanism. This paper proposes a comprehensive data-driven control framework to address these complexities without relying on analytical modeling. Our methodology comprises three core components: 1) developing a forward dynamics model (FDM) using a neural network trained on real-world experimental data to capture state transitions under varying time steps; 2) integrating this FDM into a gradient-based model predictive control (G-MPC) architecture to optimize control inputs for path following; and 3) applying imitation learning to approximate the G-MPC policy, thereby reducing the computational cost for real-time implementation. We validate the approach through simulations utilizing the identified dynamics model. The results demonstrate that the G-MPC framework achieves accurate path convergence with minimal root mean square error (RMSE), and the imitation learning controller (ILC) effectively replicates this performance. This study highlights the potential of data-driven control strategies for the precise navigation of miniature, fish-like soft robots.
}


\begin{document}

\maketitle
\thispagestyle{firstpage}

\section{Introduction}
Bio-inspired fish-like robots are rapidly gaining interest due to their exceptional efficiency and maneuverability in fluidic environments, holding vast potential for applications such as underwater exploration and environmental monitoring. Among these designs, magnetically actuated fish-like robots offer a promising solution for miniaturization; they utilize magnetic actuators to replace traditional motors, thereby eliminating the need for complex, bulky mechanical gears and joints. Consequently, this approach creates simpler, cableless, and highly agile systems, which is particularly advantageous for developing miniature robots \cite{Takada2010-cl,Takada2014-gl,Clark2015-gu,Berlinger2018-yk,Aritani2019-gg,Suzuki2019-zp,Berlinger2021-xy}.

However, despite these inherent advantages, precisely controlling these robots presents a significant technical hurdle. The difficulty primarily stems from the complex and highly nonlinear nature of fluid dynamics, including unsteady hydrodynamic forces. Crucially, this challenge is amplified when the robot features a flexible caudal fin, which introduces a complex and hysteretic relationship between the magnetic actuation input and the resulting motion. As a result, accurate modeling using conventional analytical or first-principle methods becomes extremely difficult.

Furthermore, unlike conventional robotic systems operating with fixed sampling times, the control cycle of this magnetically actuated robot is variable and action-dependent. Since the control input dictates the duration of coil excitation, the physical time step changes dynamically with each action, thereby complicating the temporal discretization required for standard predictive control models.

To overcome this challenge, existing studies have employed data-driven control techniques, such as reinforcement learning and model predictive control (MPC), to manage the intricate dynamics of fish-like robots \cite{Castano2019-ll,Zhang2020-hv}. However, most of these studies utilize conventional servo motors, and the application of data-driven control for magnetic actuators remains underexplored.

Our methodology comprises three key components: 1) learning a forward dynamics model (FDM) from real experimental data, 2) integrating the learned FDM into a gradient-based model predictive control (G-MPC) framework for path following, and 3) employing an imitation learning controller (ILC) to approximate the G-MPC policy for real-time control. In this paper, we evaluate the performance of the proposed method through simulations based on the identified dynamics model.

\section{Structure of the robot}
\begin{figure}[tbp]
  \centering
  \begin{minipage}{0.45\hsize}
    \centering
    \includegraphics[width=3.8cm]{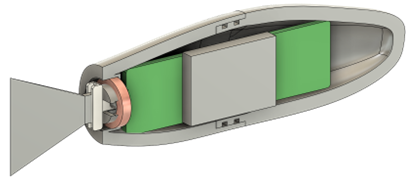}
  \end{minipage}
  \begin{minipage}{0.45\hsize}
    \centering
    \includegraphics[width=4cm]{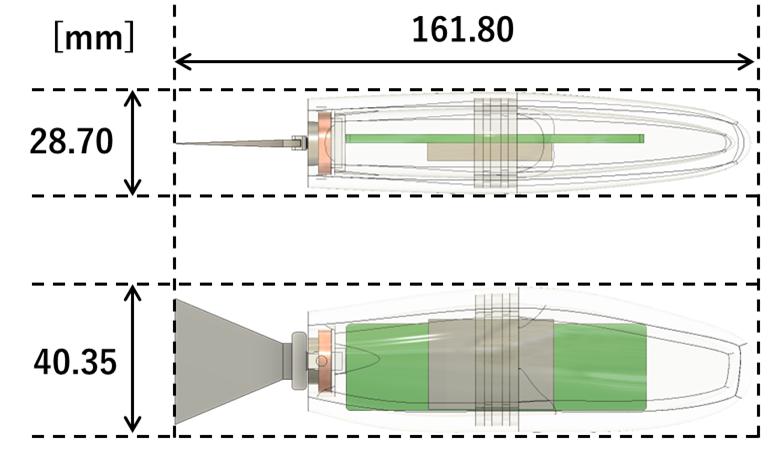}
  \end{minipage}
  \caption{Robot structure (left: cross section, right: size)}
  \label{fig:robot_structure}
\end{figure}

The robot consists of a rigid body and a flexible fin. The rigid body houses the magnet and the control electronics, while the flexible fin is made of a soft material that can bend and flex in response to the magnetic field \cite{Takada2010-cl,Takada2014-gl,Clark2015-gu,Berlinger2018-yk,Aritani2019-gg,Suzuki2019-zp,Berlinger2021-xy}.
The overall body size and cross-section of the robot are shown in Fig.~\ref{fig:robot_structure}.

\begin{figure}[tbp]
  \centering
  \includegraphics[width=6cm]{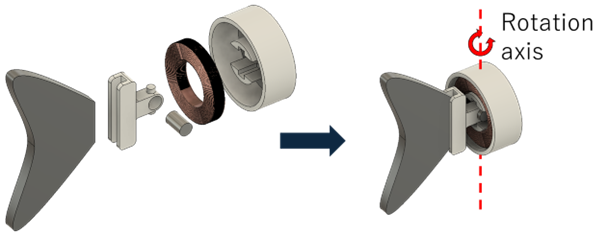}
  \caption{Magnet actuator structure}
  \label{fig:magnet_actuator}
  \vspace{5mm}
\end{figure}

\begin{figure}[tbp]
  \centering
  \begin{minipage}{0.32\hsize}
    \centering
    \includegraphics[width=2.6cm]{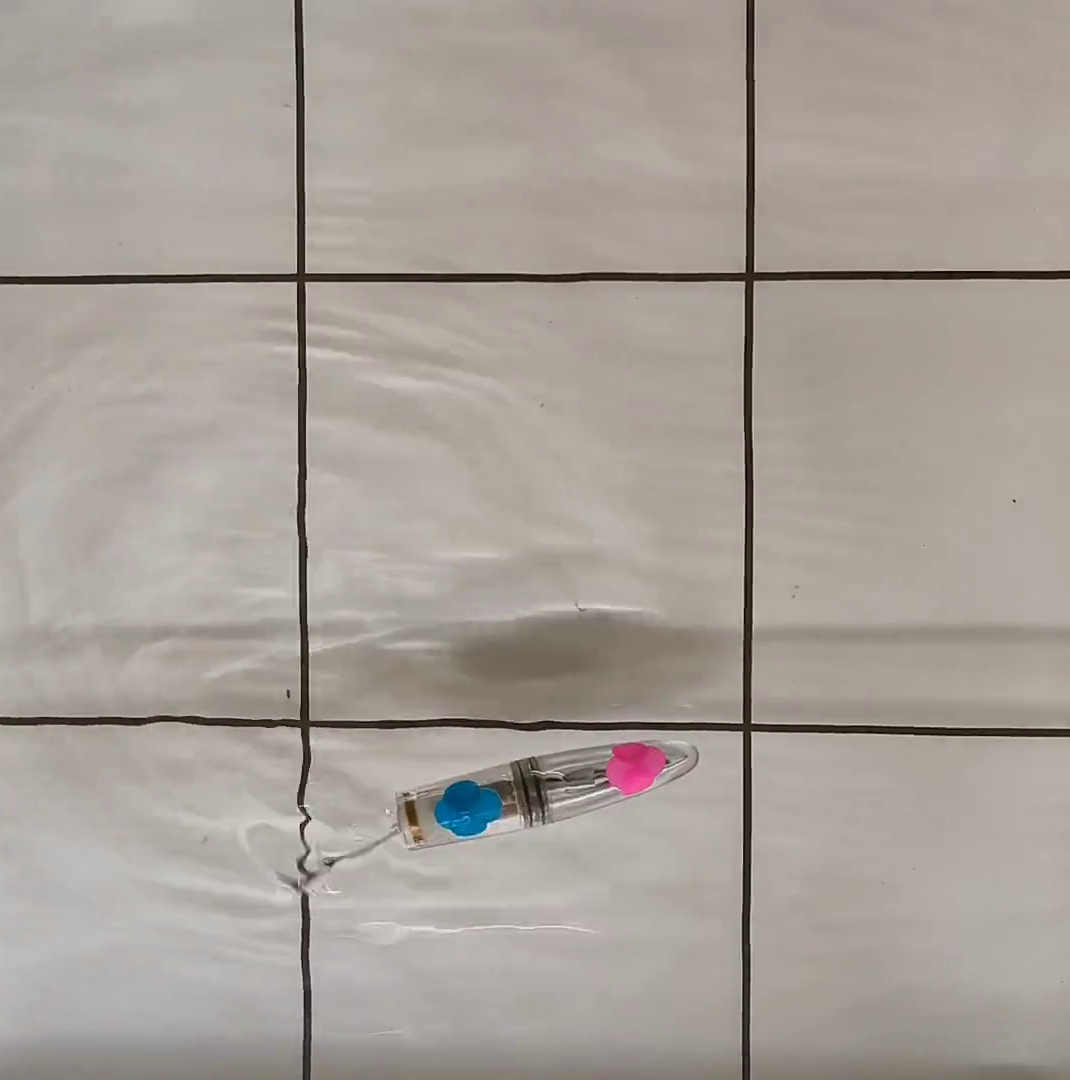}
  \end{minipage}
  \begin{minipage}{0.32\hsize}
    \centering
    \includegraphics[width=2.6cm]{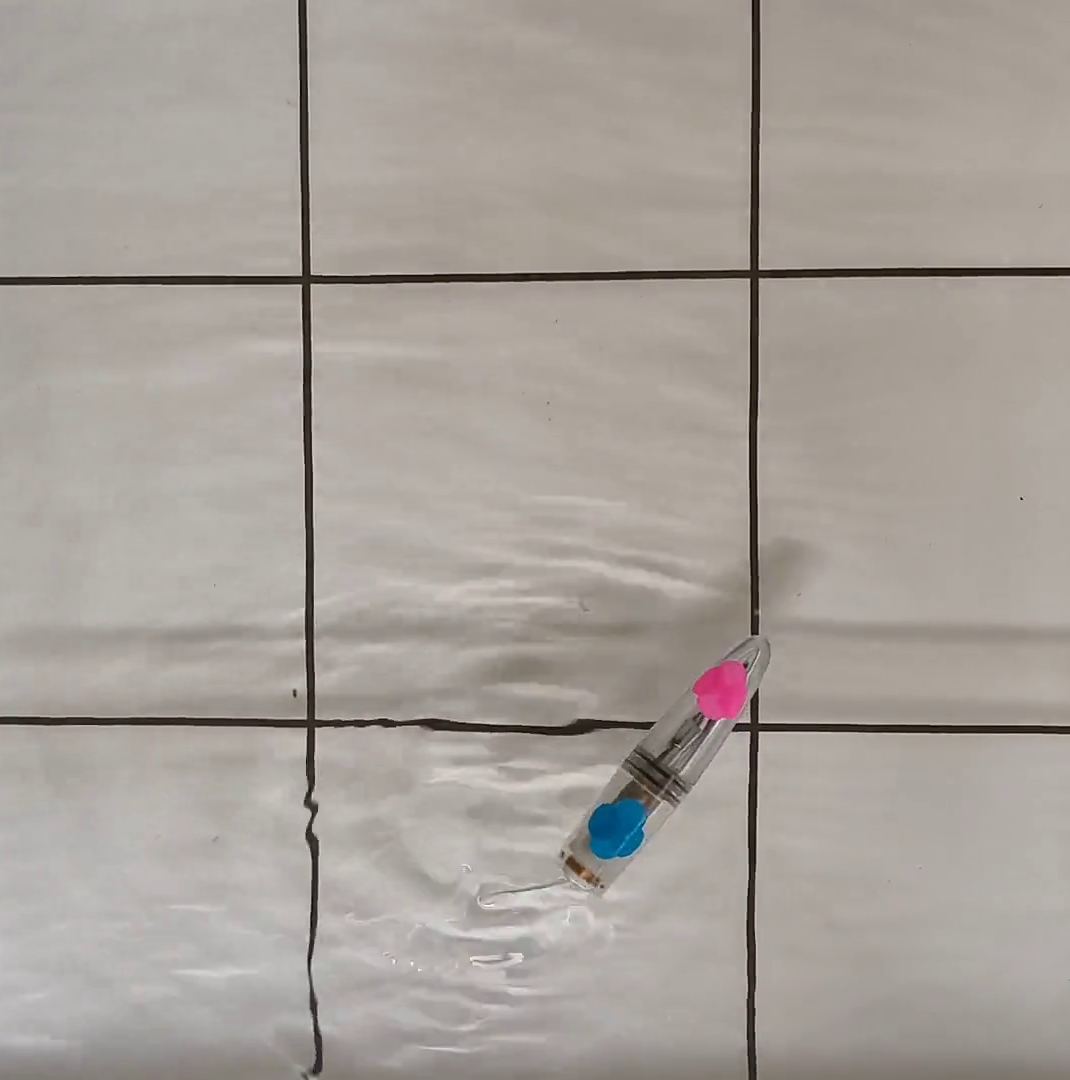}
  \end{minipage}
  \begin{minipage}{0.32\hsize}
    \centering
    \includegraphics[width=2.6cm]{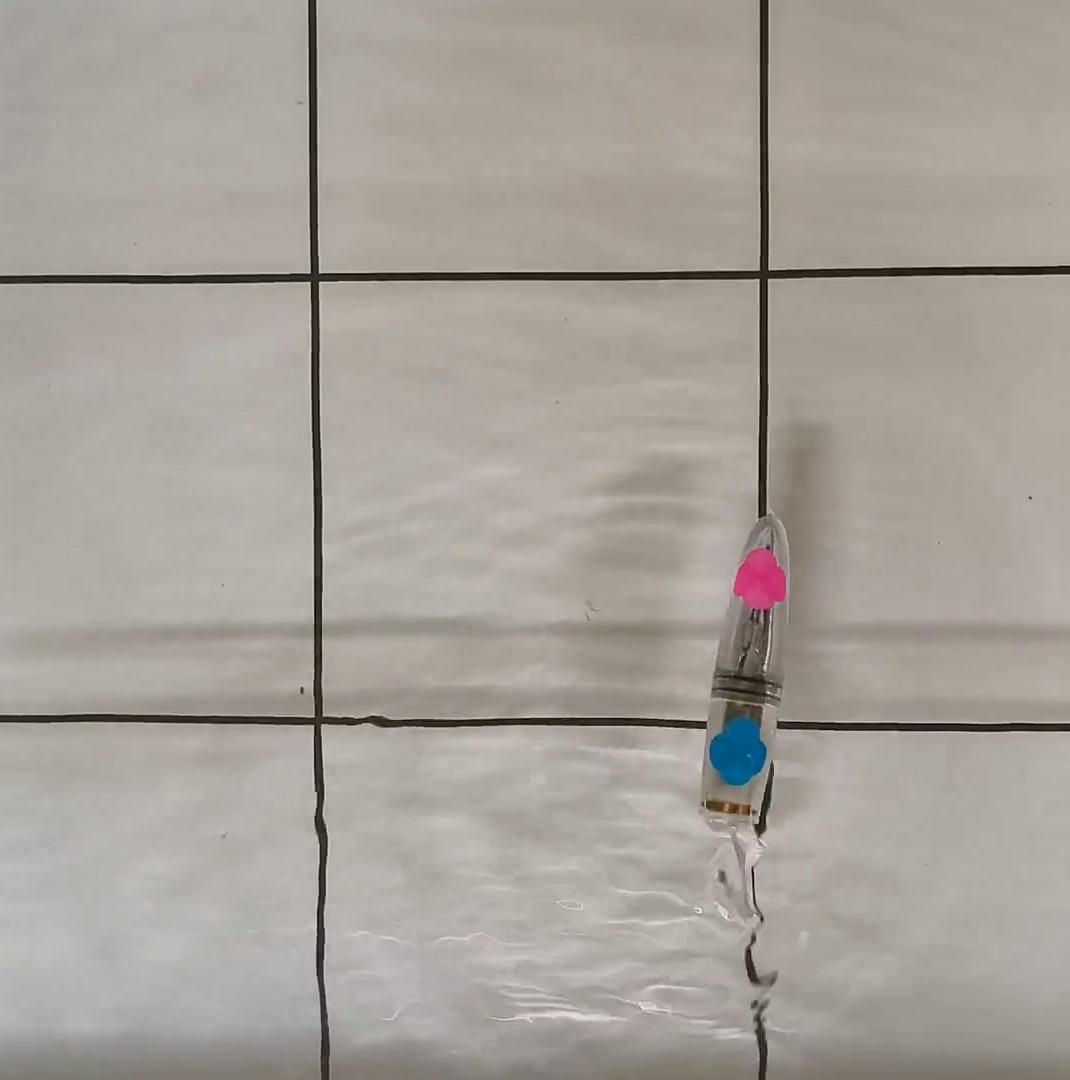}
  \end{minipage}
  \vspace{1mm}
  \caption{Robotic fish swimming sequence (1 image per second; from left to right)}
  \label{fig:real_robot_swimming_image}
  \vspace{5mm}
\end{figure}

The robot is equipped with an internal magnet actuator that generates propulsion through magnetic torque. The actuator comprises a permanent magnet and an electromagnetic coil (Fig.~\ref{fig:magnet_actuator}).
By controlling the current in the coils, the robot can manipulate the magnetic field and generate torque on the magnets, causing the flexible fin to oscillate and propel the robot forward. This architecture allows for a compact and direct actuation mechanism, eliminating the need for complex mechanical linkages. Figure~\ref{fig:real_robot_swimming_image} shows sequential images of an actual robot swimming in the water tank.

\section{Data-driven control}
\begin{figure}[tbp]
  \centering
  \includegraphics[width=8cm]{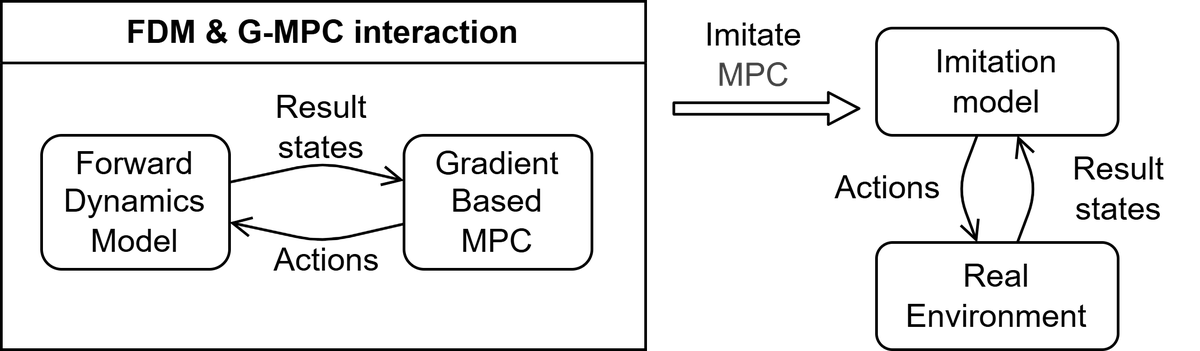}
  \caption{The architecture of control strategy}
  \label{fig:data_driven_control}
\end{figure}

The control strategy employs a data-driven approach in which imitation learning is used to approximate a G-MPC framework built upon an FDM learned from experimental data. The overall architecture of the data-driven control is shown in Fig.~\ref{fig:data_driven_control}.

\subsection{Robot-oriented local coordinate}
Let $\boldsymbol{s}^W_k=(x^W_k, y^W_k, \theta^W_k, v^W_{x,k}, v^W_{y,k}, \omega^W_k)$ be the state of the robot at step $k$ in the world coordinate system, $\boldsymbol{a}_k=(b_k, d_k)$ be the action at step $k$, and $\boldsymbol{s}^W_{k+1}$ be the next state after applying action $\boldsymbol{a}_k$ to state $\boldsymbol{s}^W_k$. Here, $(x^W_k, y^W_k)$ is the position of the robot body center in the world coordinate, $\theta^W_k$ is the angle of rigid body orientation, $(v_{x,k}^W, v_{y,k}^W)$ is the velocity, and $\omega^W_k$ is the angular velocity. The action $\boldsymbol{a}_k$ consists of two variables that control the electromagnetic actuator. The action variables $b_k$ and $d_k$ represent the duration of current application (on-time) for the left and right coils, respectively, measured in milliseconds. Unlike conventional discrete-time control with a fixed time step $\Delta t$, the physical duration of each control step $k$ depends on the chosen action, defined roughly as $\Delta t_k \approx b_k + d_k$.
In our experiments, these values typically range from 200 to 1100, allowing for varied propulsion patterns. Consequently, the FDM is trained to predict the state transition $\boldsymbol{s}_{k+1}$ occurring after the specific duration dictated by $\boldsymbol{a}_k$, implicitly learning the time-dependency of the dynamics. Before being fed into the neural network, these action values are normalized to a range close to $[0, 1]$ based on their operational limits (min: 200, max: 900 with reference to data collection configuration) to ensure training stability.

\begin{figure}[tbp]
  \centering
  \begin{minipage}{0.45\hsize}
    \centering
    \includegraphics[width=4cm]{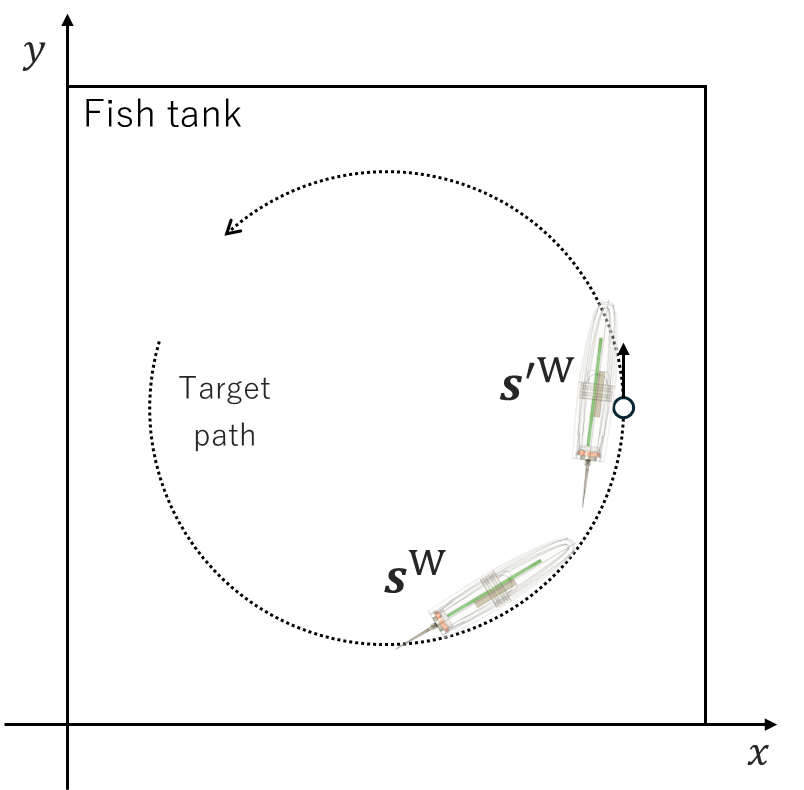}
  \end{minipage}
  \begin{minipage}{0.45\hsize}
    \centering
    \includegraphics[width=3.5cm]{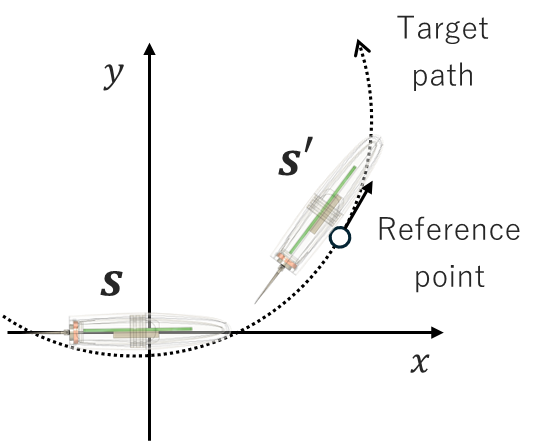}
  \end{minipage}
  \vspace{1mm}
  \caption{World coordinate (left) and robot-oriented local coordinate (right)}
  \label{fig:about_coordinate}
\end{figure}

We define a robot-oriented local coordinate system as follows (Fig.~\ref{fig:about_coordinate}); we refer to this as the robot coordinate. The x-axis is aligned with $\theta^W_k$, and the y-axis is perpendicular to it. The origin is set at $(x^W_k, y^W_k)$. The state in the robot coordinate is defined as $\boldsymbol{s}_k=(0, 0, 0,v_{x,k}, v_{y,k}, \omega_k)$, where $v_{x,k}$ and $v_{y,k}$ are the velocities in the x and y directions of the robot coordinate, and $\omega_k$ is the angular velocity. The three zeros indicate that the position is relative to the origin of the robot coordinate, and this notation is used to align the dimensions with those of $\boldsymbol{s}_{k+1}$ described as follows. The next state in the robot coordinate is denoted as $\boldsymbol{s}_{k+1}=(x_{k+1}, y_{k+1}, \theta_{k+1}, v_{x,k+1}, v_{y,k+1}, \omega_{k+1})$, where $(x_{k+1}, y_{k+1})$ is the position of the robot body center in the robot coordinate, $\theta_{k+1}$ is the angle of rigid body orientation, $(v_{x,k+1}, v_{y,k+1})$ is the velocity, and $\omega_{k+1}$ is the angular velocity.

\subsection{Forward dynamics model (FDM)}
We employ a neural network, specifically a multilayer perceptron (MLP), to learn the FDM. The output of the FDM is defined as $\hat{\boldsymbol{s}}_{k+1} = \text{FDM}(\boldsymbol{s}_k, \boldsymbol{a}_k)$, where $\hat{\boldsymbol{s}}_{k+1}$ is the predicted state at step $k+1$. The FDM serves as a function approximator for robot state transitions in water and is trained using supervised learning on a dataset $\{(\boldsymbol{s}_k, \boldsymbol{a}_k, \boldsymbol{s}_{k+1})\}_k$ collected from real-world experiments.

The architecture of the neural network consists of three fully connected layers with ReLU activation functions. The input layer takes the concatenated vector of $\boldsymbol{s}_k$ and $\boldsymbol{a}_k$, and the output layer produces the predicted next state $\hat{\boldsymbol{s}}_{k+1}$.
The loss function is defined as the mean squared error between the predicted next state $\hat{\boldsymbol{s}}_{k+1}$ and the actual next state $\boldsymbol{s}_{k+1}$. During training, we use the Adam optimizer.

For data collection, we conduct experiments in the setup detailed later in Sec.~\ref{sec:datacollection}, applying various control inputs to the electromagnetic actuator and recording the resulting state transitions. Concretely, we vary the on-times $b_k$ and $d_k$ of the electromagnetic actuator within specified ranges to ensure diverse state transitions, and capture before and after states at each control step.

The trained FDM is capable of predicting the next state accurately owing to its nonlinear network structure.

\subsection{Gradient-based MPC (G-MPC)}
In this study, to address the path-following task, we formulate the problem within a model predictive control (MPC) framework. While FDM represents the forward dynamics $\hat{\boldsymbol{s}}_{k+1} = \text{FDM}(\boldsymbol{s}_k, \boldsymbol{a}_k)$, G-MPC serves as an inverse dynamics solver that computes the optimal action sequence given the current state and a target path.

Formally, we define G-MPC as a function that maps the current state $\boldsymbol{s}_k$ to a sequence of optimal control inputs $\bm{\mathcal{A}}^*_k$, given a target path (discretized parametric curve) $\bm{\mathcal{P}}^* = \{ \boldsymbol{p}_i\}_i$:
\begin{equation}
\bm{\mathcal{A}}^*_k = \text{G-MPC}(\boldsymbol{s}_k, \bm{\mathcal{P}}^*) = \operatorname{arg\,min}_{\bm{\mathcal{A}}_k} J(\boldsymbol{s}_k, \bm{\mathcal{A}}_k, \bm{\mathcal{P}}^*)
\end{equation}
where $\bm{\mathcal{A}}_k = \{\boldsymbol{a}_k, \boldsymbol{a}_{k+1}, \ldots, \boldsymbol{a}_{k+H-1}\}$ is a sequence of control inputs over the prediction horizon $H$.
As in standard MPC, only the first optimized input $\boldsymbol{a}^*_k$ is used for control.
Note that, as a reference to achieve smooth control, the target path consists of $\boldsymbol{p}_i = (x_i, y_i, \theta_i)$, where $\theta_i$ is the angle of the tangent vector $(\dot{x}, \dot{y})$ of the original parametric curve at $(x_i, y_i)$.

G-MPC follows the MPC receding horizon strategy of optimizing a sequence of control inputs over a prediction horizon $H$ rather than directly inferring a single action.
Specifically, the objective function $J$ is given by the cumulative cost over the horizon:
\begin{equation}
J = \sum_{j=0}^{H-1} \text{Cost}(\hat{\boldsymbol{s}}_{k+j+1}, \boldsymbol{p}^*_{\text{ref}, k+j+1}),
\end{equation}
and is minimized subject to the dynamics constraints $\hat{\boldsymbol{s}}_{k+j+1} = \text{FDM}(\hat{\boldsymbol{s}}_{k+j}, \boldsymbol{a}_{k+j})$.
This optimization is performed via gradient descent, backpropagating the gradients through the differentiable FDM. The G-MPC algorithm is summarized in Algorithm~\ref{alg:gmpc}.

The reference point $\boldsymbol{p}^*_{\text{ref}, k+j+1}$ is selected from the target path $\bm{\mathcal{P}}^*$ using a specific search strategy to handle path geometries. First, we identify a ``nearest point'' on the path that minimizes a weighted error metric combining position and heading deviations.
Specifically, for every point $\boldsymbol{p}_i \in \bm{\mathcal{P}}^*$, we compute the error $E(i) = \|\boldsymbol{p}_{i,\text{xy}} - \hat{\boldsymbol{s}}_{\text{xy}}\|^2 + w_h (\boldsymbol{p}_{i,\theta} - \hat{s}_{\theta})^2$, where $(\cdot)_{\text{xy}}$ and $(\cdot)_{\theta}$ denote the vector components corresponding to the subscripts and $w_h$ is the heading weight.
We then select the nearest point
$\boldsymbol{p}_{i'}$ with $i' = \arg\min_i E(i)$.
Subsequently, searching forward from this index (i.e., along the future path index, $i \geq i'$), we select the candidate point whose distance from the current position $\hat{\boldsymbol{s}}_\text{xy}$ is closest to the defined look-ahead distance $L_{\text{lookahead}}$.

\begin{algorithm}[tbp]
  \caption{Gradient-Based Model Predictive Control (G-MPC)}
  \label{alg:gmpc}
  \begin{algorithmic}[1]

    \Statex \textbf{Parameters:}
      \Statex \quad Prediction horizon $H$, Optimization iterations $N$
      \Statex \quad Look-ahead distance $L_{\text{lookahead}}$, Heading weight $w_h$
      \Statex \quad Weight matrix $W$ (diagonal)
    \Statex \textbf{Functions:}
      \Statex \quad $\textsc{GetRef}(\hat{\boldsymbol{s}}, \bm{\mathcal{P}}^*)$:
      \Statex \qquad 1. Find nearest point index $i' = \arg\min_i E(i)$. %
      \Statex \qquad 2. Find reference point index
      \Statex \qquad\qquad  $i^* = \arg\min_{i\geq i'} |\|\boldsymbol{p}_{i,\text{xy}} - \hat{\boldsymbol{s}}_\text{xy}\| - L_{\text{lookahead}}|$
      \Statex \qquad 3. Return $\boldsymbol{p}_{i^*}$.
      \Statex \quad $\textsc{Cost}(\hat{\boldsymbol{s}}, \boldsymbol{p}^*_{\text{ref}}) = (\hat{\boldsymbol{s}}_{xy\theta} - \boldsymbol{p}^*_{\text{ref}})^\top W (\hat{\boldsymbol{s}}_{xy\theta} - \boldsymbol{p}^*_{\text{ref}})$

      \Statex \hrulefill

      \Statex \textbf{Input:}
      \Statex \quad Current state $\boldsymbol{s}_k$
      \Statex \quad Target path $\bm{\mathcal{P}}^*$
      \Statex \quad Initial action sequence $\bm{\mathcal{A}}_k = \{\boldsymbol{a}_k, \ldots, \boldsymbol{a}_{k+H-1}\}$
      \Statex \textbf{Output:}
      \Statex \quad Optimal control inputs $\bm{\mathcal{A}}_k$

      \Statex \hrulefill

      \For{$i = 1$ to $N$}  \Comment{Optimization iteration loop}
          \State $\hat{\boldsymbol{s}}_{k} = \boldsymbol{s}_{k}$ \Comment{Initialize prediction with current state}
          \State $J_{\text{total}} = 0$
          \For{$j = k$ to $k+H-1$}  \Comment{Horizon loop}
              \State $\boldsymbol{p}^*_{\text{ref}, j+1} = \textsc{GetRef}(\hat{\boldsymbol{s}}_{j}, \bm{\mathcal{P}}^*)$
              \State $\hat{\boldsymbol{s}}_{j+1} = \text{FDM}(\hat{\boldsymbol{s}}_{j}, \boldsymbol{a}_{j})$
              \State $J_{\text{total}} \leftarrow J_{\text{total}} + \textsc{Cost}(\hat{\boldsymbol{s}}_{j+1}, \boldsymbol{p}^*_{\text{ref}, j+1})$
          \EndFor
          \State Compute gradients $\nabla_{\bm{\mathcal{A}}_k} J_{\text{total}}$
          \State Update action sequence $\bm{\mathcal{A}}_k$ using Adam optimizer
      \EndFor
      \State \textbf{return} $\bm{\mathcal{A}}_k$ \Comment{Return optimized control inputs}
  \end{algorithmic}
  \end{algorithm}

\subsection{Imitation learning controller (ILC)}
Generally, applying G-MPC in real-time applications can be computationally intensive due to the iterative optimization process required at each time step. To address this problem, we employ an imitation learning controller (ILC), defined as $\boldsymbol{a}_k = \text{ILC}(\boldsymbol{s}_k, \boldsymbol{p}^*_{\text{ref},k+1})$, to generate control inputs with a single forward-pass inference through a neural network (an MLP). Specifically, we first create a dataset of near-optimal control inputs $\{ (\boldsymbol{s}_k, \boldsymbol{p}_{\text{ref},k+1}, \boldsymbol{a}^*_{k}) \}_k$ by running the G-MPC offline over a variety of scenarios and recording the resulting semi-optimal actions for given states and reference points. Then, we train the ILC to learn the mapping from states and reference points to these control inputs using supervised learning on this dataset.

\section{Experiments}
\subsection{Data collection setup}
\label{sec:datacollection}
We evaluate the performance of the integrated FDM and G-MPC approach through simulations based on the experimentally derived dynamics model.

\begin{figure}[tbp]
\centering
  \vspace{5mm}
\includegraphics[width=8cm]{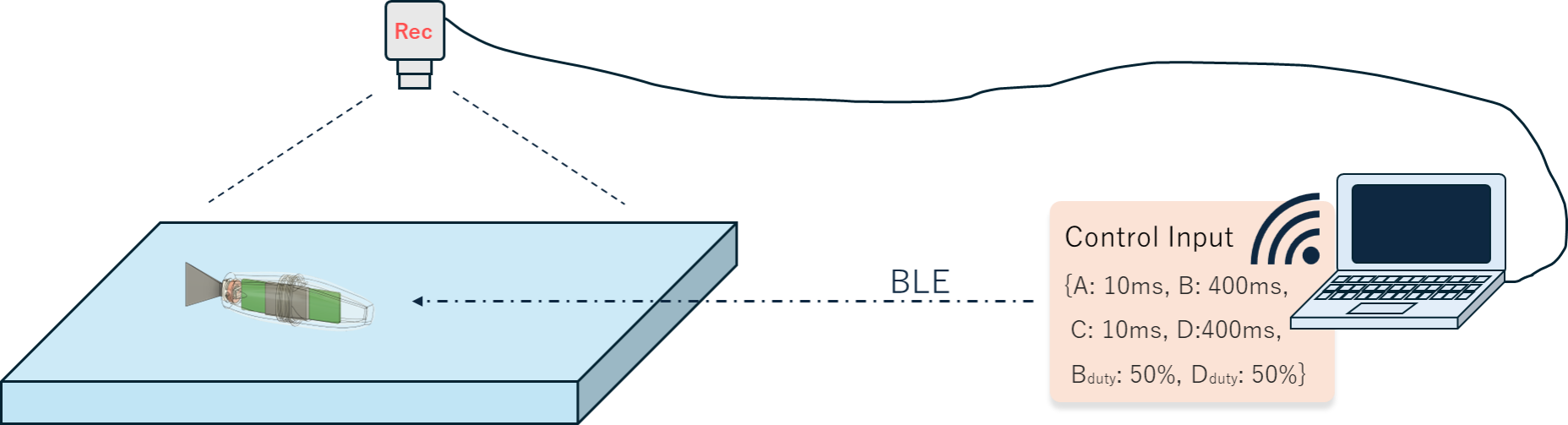}
\caption{Experimental setup}
\label{fig:experimental_setup}
\end{figure}

In this experiment, we use a water tank with both length and width of 600~mm, and a water depth of about 100~mm (Fig.~\ref{fig:experimental_setup}). The robot is placed in the tank, and its motion is tracked using an overhead camera system. The camera captures the position and orientation of the robot at approximately 30~Hz. The control inputs to the electromagnetic actuator are generated using an external computer that runs the imitation model and sends the commands to the robot via wireless communication.

To evaluate the proposed method, we conducted simulations to assess the robot's path-following performance using the data-driven control approach. In the simulation, we utilized the learned FDM to simulate the robot's motion in response to control inputs generated by the G-MPC and imitation learning controller. Both tests were evaluated through a root mean square error (RMSE) metric, which quantifies the average deviation between the reference point on the target path and the robot's actual position after each control step.

\subsection{G-MPC performance}
To demonstrate the effectiveness of the G-MPC framework, we present path-following results obtained from simulation utilizing the learned FDM. The parameters for the FDM and G-MPC used in this evaluation are listed in Tables~\ref{tab:fdm_parameters} and \ref{tab:gmpc_parameters}, respectively. For the FDM training, we collected a dataset comprising approximately 300 (state, action, next state) tuples from real-world experiments by applying various control inputs to the electromagnetic actuator and recording the resulting state transitions. The target path was generated using a Bezier curve to create a smooth 90-degree right turn. The robot was initialized at three different starting positions: above, on, and below the target path. The other state variables are set to 0.

\begin{figure}[bp]
  \centering
  \begin{minipage}{0.49\linewidth}
    \centering
    \includegraphics[width=\linewidth]{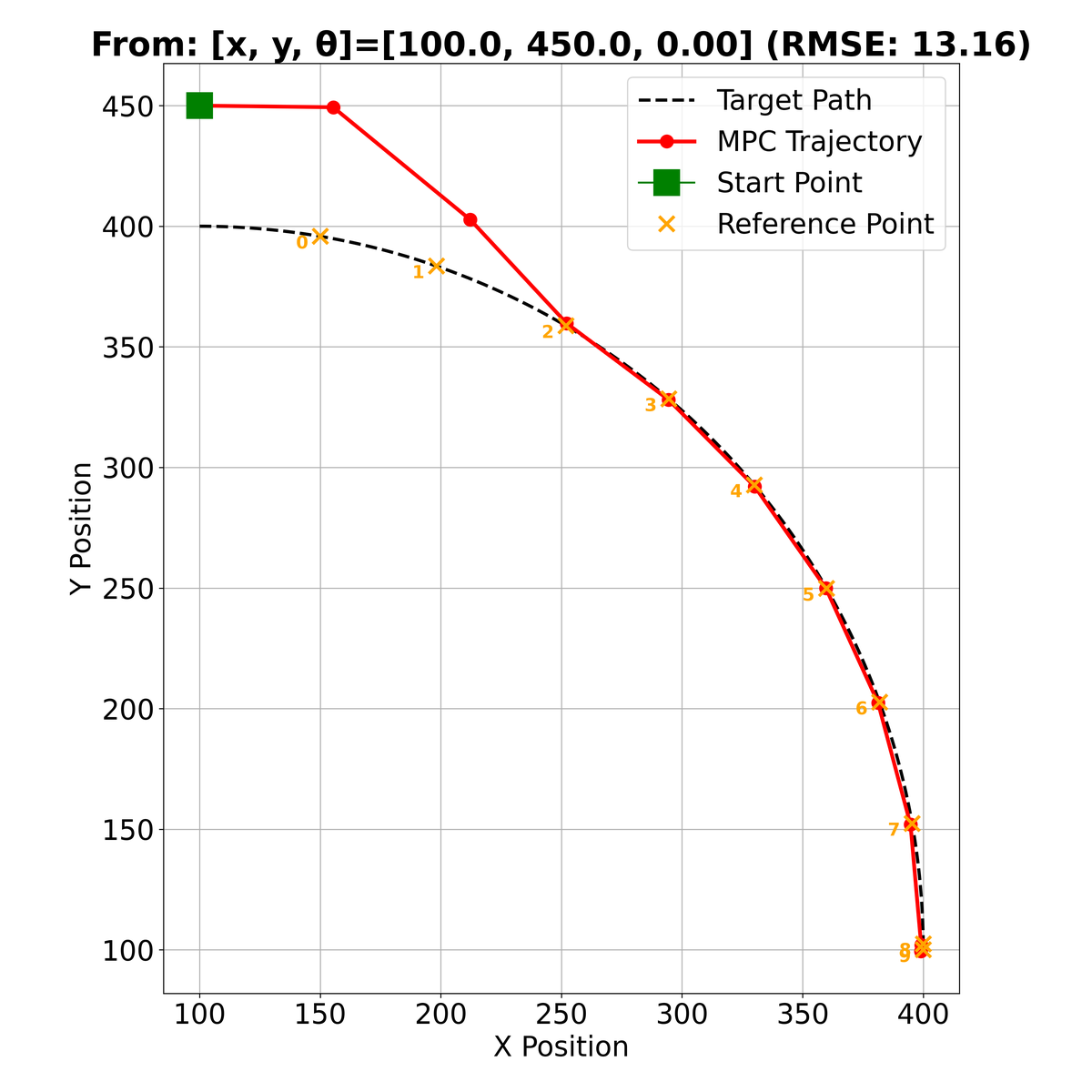}
  \end{minipage}
  \hfill
  \begin{minipage}{0.49\linewidth}
    \centering
    \includegraphics[width=\linewidth]{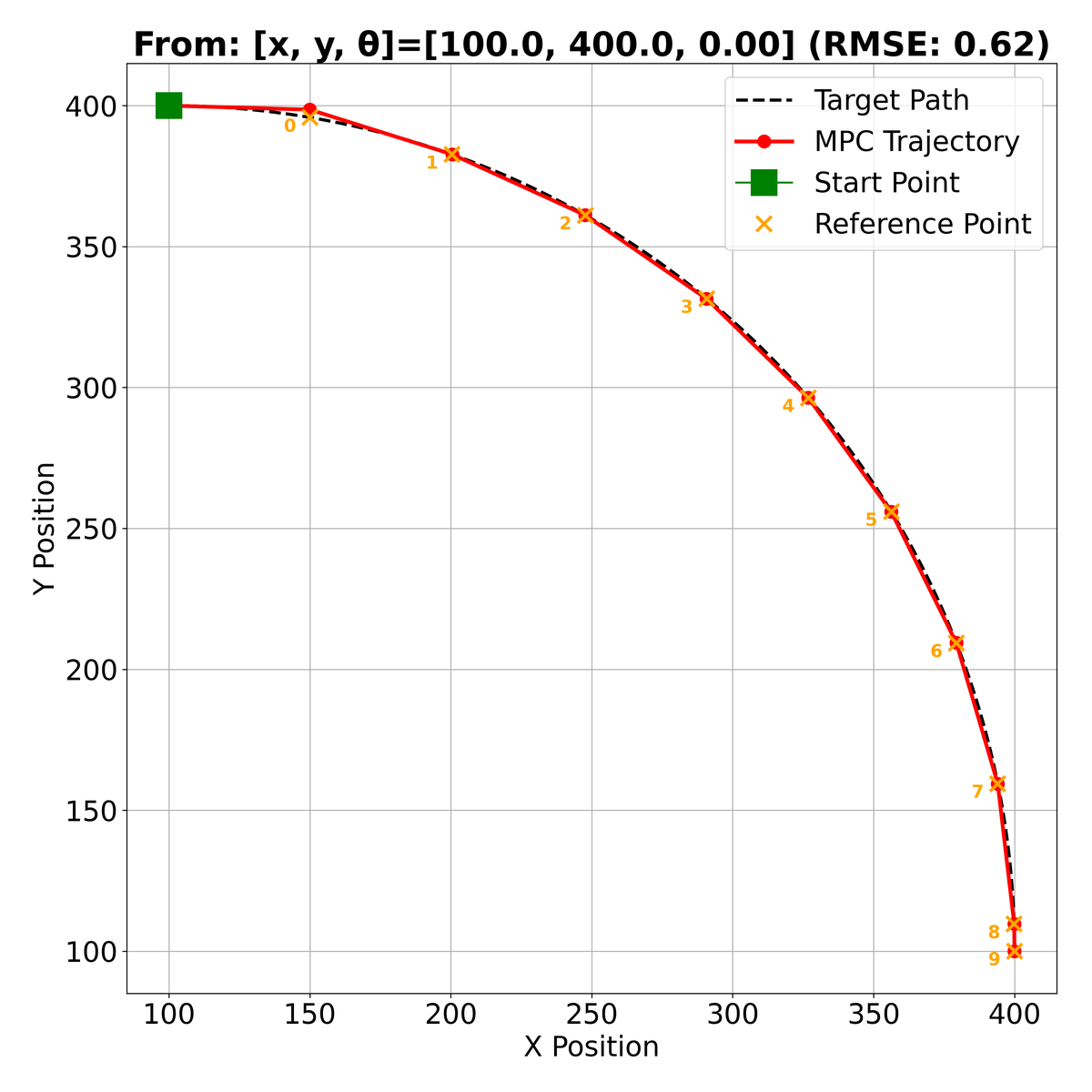}
  \end{minipage}

  \vspace{1em}

  \begin{minipage}{0.49\linewidth}
    \centering
    \includegraphics[width=\linewidth]{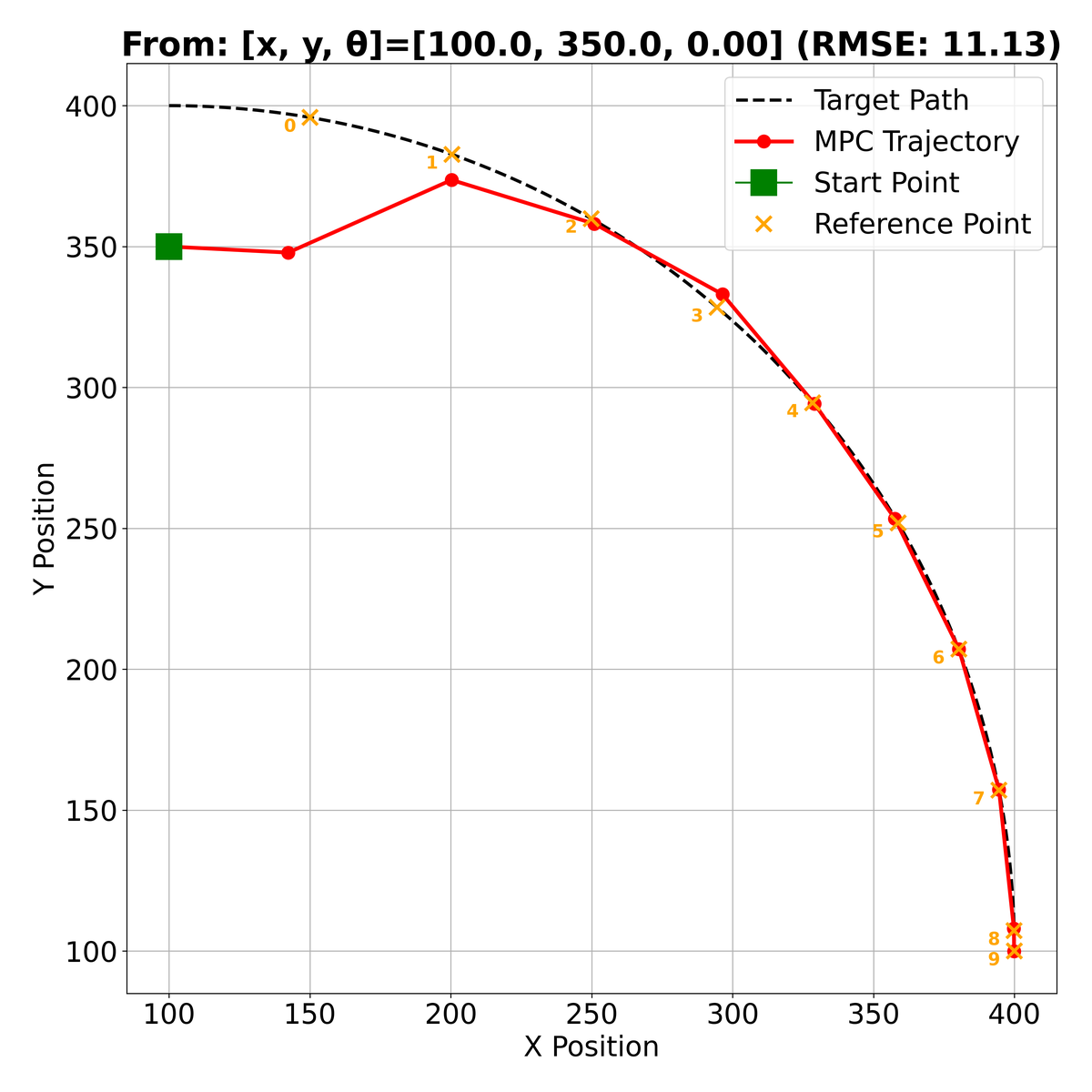}
  \end{minipage}

  \caption{G-MPC path-following results from different start positions (top left: above path, top right: on path, bottom: below path). Red line: connection of the output positions after each control step. Black dashed line: target path. Orange cross: reference point each control step. Green square: start position.}
  \label{fig:gmpc_results}
\end{figure}

The results of the path following using G-MPC are shown in Fig.~\ref{fig:gmpc_results}. In the cases of starting above and below the path, the robot successfully converges to the target path and follows it closely after some oscillations, with RMSE values of 13.16 mm and 11.13 mm, respectively. In the case of starting on the path, the robot maintains its position along the path with minimal deviation, resulting in an RMSE of 0.62 mm. These results demonstrate the effectiveness of the G-MPC framework in achieving accurate path following using the learned FDM.

\begin{table}[tbp]
  \centering
  \caption{FDM parameters}
  \label{tab:fdm_parameters}
  \begin{tabular}{|l|c|}
  \hline
  Parameter & Value \\
  \hline
  Number of layers & 3 \\
  \hline
  Number of units per layer & 8 \\
  \hline
  Activation function & ReLU \\
  \hline
  Optimizer & AdamW \\
  \hline
  Learning rate & 0.001 \\
  \hline
  Batch size & 8 \\
  \hline
  Number of epochs & 100 \\
  \hline
  \end{tabular}
  \end{table}

  \begin{table}[tbp]
  \centering
  \caption{G-MPC parameters}
  \label{tab:gmpc_parameters}
  \begin{tabular}{|l|c|}
  \hline
  Parameter & Value \\
  \hline
  Prediction horizon ($H$) & 10 \\
  \hline
  Optimization iterations ($N$) & 1000 \\
  \hline
  Look-ahead distance ($L_{\text{lookahead}}$) & 50.0 mm \\
  \hline
  Learning rate & 0.005 \\
  \hline
  Action dimension & 2 \\
  \hline
  Loss weight (x, y, yaw) & (5.0, 5.0, 1.0) \\
  \hline
  Optimizer & AdamW \\
  \hline
  \end{tabular}
  \end{table}

\subsection{ILC performance}
\begin{figure}[tbp]
  \centering
  \includegraphics[width=6cm]{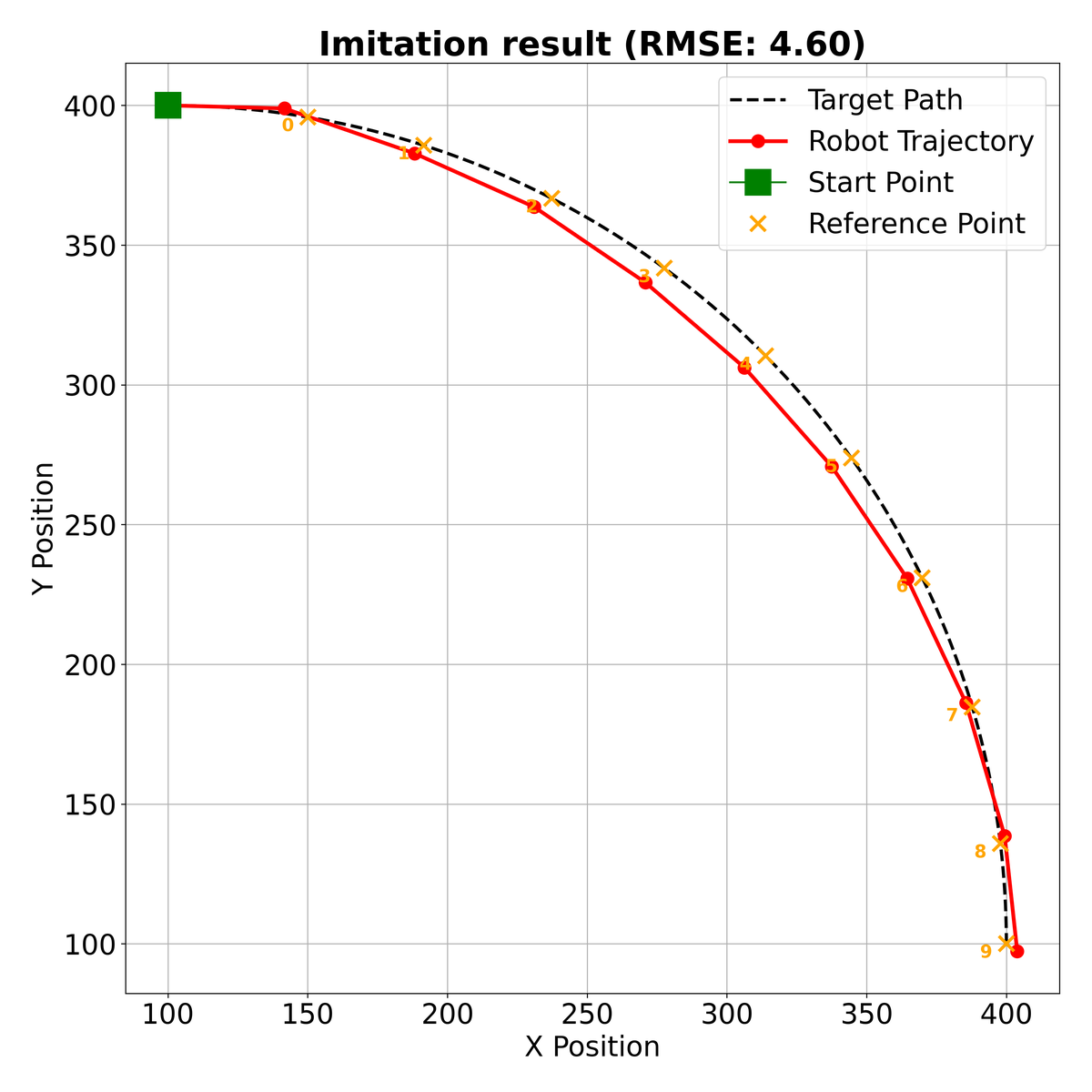}
  \caption{ILC path-following result}
  \label{fig:imitation_learning_result}
  \vspace{5mm}
\end{figure}

Next, we evaluated the ILC performance in simulation, following a procedure similar to the G-MPC evaluation. The parameters of the imitation learning are shown in Table~\ref{tab:imitation_learning_parameters}. To generate the training dataset, we executed G-MPC from various initial positions and collected the resulting state-action pairs on the same curved path used in the G-MPC evaluation. Specifically, the robot was initialized at x-coordinates ranging from 50 to 150~mm and y-coordinates from 350 to 450~mm in 50~mm intervals, combined with yaw angles of $-\pi/18$, $0$, and $\pi/18$ radians. An example of the path-following results using ILC is shown in Fig.~\ref{fig:imitation_learning_result}.
The robot successfully follows the target path with an RMSE of 4.60 mm, demonstrating that the ILC effectively approximates the G-MPC behavior for path-following tasks.

\begin{table}[tbp]
\centering
\caption{Imitation learning parameters}
\label{tab:imitation_learning_parameters}
\begin{tabular}{|l|c|}
\hline
Parameter & Value \\
\hline
Number of layers & 2 \\
\hline
Number of units per layer & 8 \\
\hline
Learning rate & 0.1 \\
\hline
Batch size & 32 \\
\hline
Number of epochs & 1000 \\
\hline
Optimizer & AdamW \\
\hline
\end{tabular}
\end{table}

\section{Conclusion}
This research presents a data-driven control approach for a magnetically actuated fish-like robot with a flexible fin. By integrating a learned forward dynamics model into a gradient-based model predictive control framework and employing imitation learning, we developed an effective path-following control strategy. The simulation results demonstrated the capability of the proposed method to achieve accurate path following, with low RMSE values under various initial conditions.

In this study, we primarily focused on validating the overall control framework through simulations based on the learned dynamics model. While the results are promising, further experimental validation on a physical robot is necessary to fully assess the accuracy and robustness of the forward dynamics model and the control strategy in real-world conditions.

Future work will focus on real robot experiments to validate the approach from simpler paths to more complex paths, as well as exploring robustness against disturbances and uncertainties in the environment.

\section*{acknowledgement}
This work was supported by JSPS KAKENHI Grant Number JP21H05302.

\end{document}